\documentclass[fleqn]{SCYE}

\usepackage{multirow}
\usepackage{enumitem}

\usepackage{nccmath}

\begin{document}

\ArticleType{REVIEW}
\Year{2020}
\Month{March}
\Vol{01}
\No{1}
\DOI{10.1007/s11432-016-0037-0}
\BeginPage{100} 
\EndPage{120}
\ReceiveDate{March 20, 2020}
\AcceptDate{April 20, 2020}
\OnlineDate{May 20, 2020}

\title{Recent Advances and Challenges in \\Task-oriented Dialog Systems}{Recent Advances and Challenges in Task-oriented Dialog Systems}

\author[]{Zheng Zhang}{}
\author[]{Ryuichi Takanobu}{}
\author[]{Qi Zhu}{}
\author[]{Minlie Huang}{{aihuang@tsinghua.edu.cn}}
\author[]{Xiaoyan Zhu}{}

\AuthorMark{Zhang Z, et al.}

\AuthorCitation{Zhang Z, Takanobu R, Zhu Q, Huang M, Zhu X}

\address[]{Dept. of Computer Science \& Technology, Tsinghua University, Beijing {\rm 100084}, China}
\address[]{Institute for Artificial Intelligence, Tsinghua University (THUAI), Beijing {\rm 100084}, China}
\address[]{Beijing National Research Center for Information Science \& Technology, Beijing {\rm 100084}, China}


\abstract{
Due to the significance and value in human-computer interaction and natural language processing, task-oriented dialog systems are attracting more and more attention in both academic and industrial communities. 
In this paper, we survey recent advances and challenges in task-oriented dialog systems. We also discuss three critical topics for task-oriented dialog systems: (1) improving data efficiency to facilitate dialog modeling in low-resource settings, (2) modeling multi-turn dynamics for dialog policy learning to achieve better task-completion performance, and (3) integrating domain ontology knowledge into the dialog model.
Besides, we review the recent progresses in dialog evaluation and some widely-used corpora.
We believe that this survey, though incomplete, can shed a light on future research in task-oriented dialog systems.}


\maketitle



\begin{multicols}{2}

\section{Introduction}

Building task-oriented (also referred to as goal-oriented) dialog systems has become a hot topic in the research community and the industry. A task-oriented dialog system aims to assist the user in completing certain tasks in a specific domain, such as restaurant booking, weather query, and flight booking, which makes it valuable for real-world business.
Compared to open-domain dialog systems where the major goal is to maximize user engagement \cite{huang2019challenges},
task-oriented dialog systems are more targeting at accomplishing some specific tasks in one or multiple domains \cite{chen2017survey}. Typically, task-oriented dialog systems are built on top of a structured ontology, which defines the domain knowledge of the tasks.


Existing studies on task-oriented dialog systems can be broadly classified into two categories: pipeline and end-to-end methods. In the pipeline methods, the entire system is divided into several modules, including natural language understanding (NLU), dialog state tracking (DST), dialog policy (Policy) and natural language generation (NLG). There are also some other combination modes, such as word-level DST \cite{mrkvsic2017neural,wu2019transferable} (coupling NLU and DST) and word-level policy \cite{zhao2019rethinking,chen2019semantically} (coupling Policy and NLG). 
While end-to-end methods build the system using a single model, which directly takes a natural language context as input and outputs a natural language response as well.

Building pipeline system often requires large-scale labeled dialog data to train each component. The modular structure makes the system more interpretable and stable than end-to-end counterparts. Therefore, most real-world commercial systems are built in this manner. End-to-end systems require less annotations, making it more easily to build. However, the end-to-end structure makes it a black box, which is more uncontrollable \cite{gao2018neural}.

\begin{figure*}[!t]
	\includegraphics[width=1.0\linewidth]{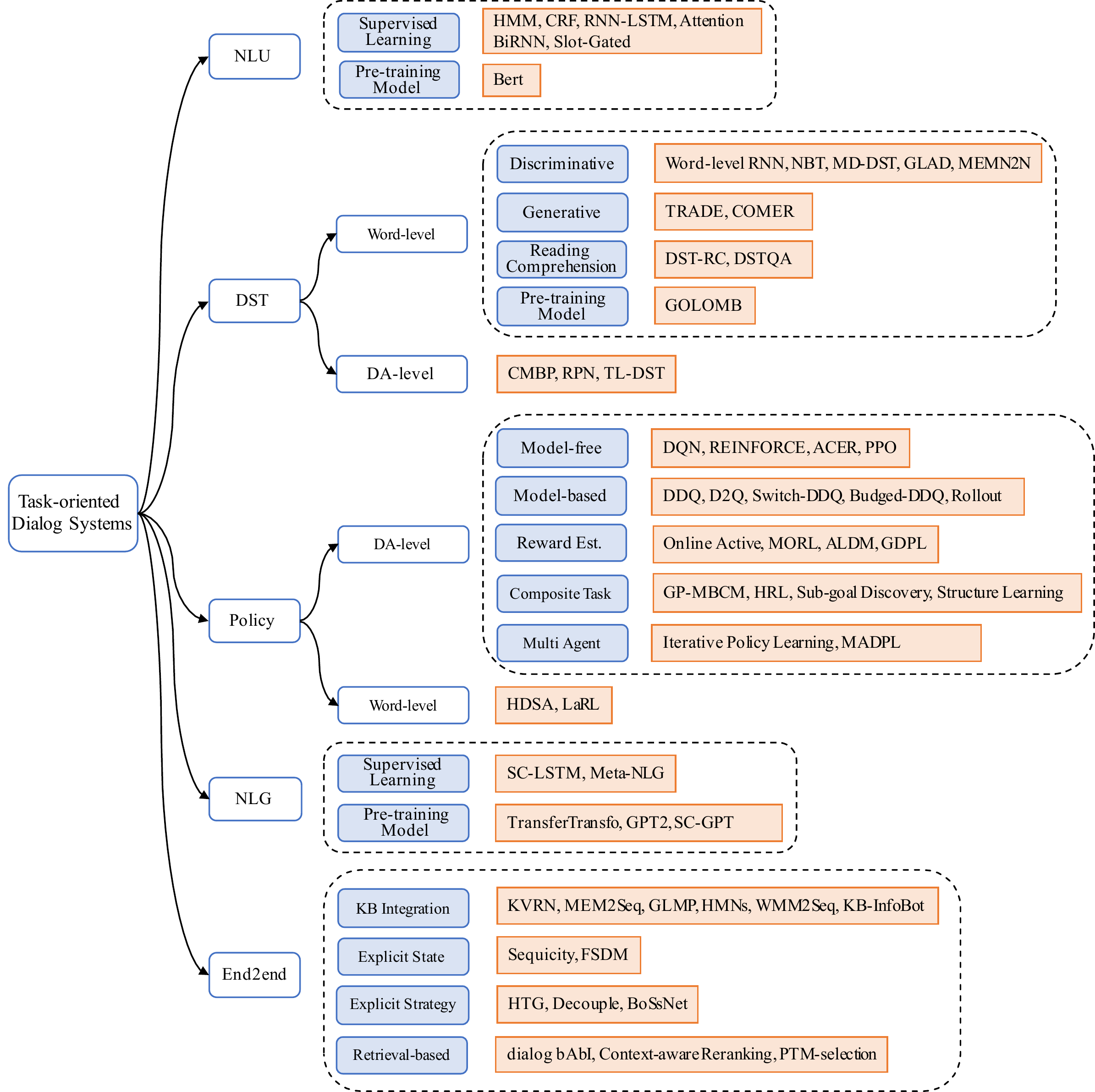}
	\caption{Taxonomy of task-oriented dialog systems.}
	\label{fig:taxonomy}
\end{figure*}

To make a clear review of existing studies, we build a taxonomy for task-oriented dialog systems. As illustrated in Figure \ref{fig:taxonomy}, for each individual component in pipeline and end-to-end methods, we list several key issues within which typical works are presented.

In pipeline methods, recent studies focus more on the dialog state tracking and dialog policy components, which are also called \textit{Dialog Management}. This is because both NLU and NLG components are standalone language processing tasks, which are less interweaved to the other tasks in dialog systems.
Based on the domain ontology, the DST task can be seen as a classification task by predicting the value of each slot. However, when the training data are not sufficient, such classification-based methods can suffer from the out-of-vocabulary (OOV) problem and can not be directly generalized to new domains. The dialog policy learning task is often considered as a reinforcement learning task. Nevertheless, different from other well-known RL tasks, such as playing video games \cite{mnih2013playing} and Go \cite{silver2016mastering}, the training of dialog policy requires real humans to serve as the environment, which is very costly. Furthermore, most existing methods used manually defined rewards, such as task-completion rate and session turn number, which cannot reliably evaluate the performance of a system.

For end-to-end methods, the data-hungry nature of the vanilla sequence-to-sequence model makes it difficult to learn the sophisticated slot filling mechanism in task-oriented dialog systems with a limited amount of domain-specific data. The knowledge base query issue requires the model to generate an intermediate query besides the encoder and the decoder, which is not straightforward. Another drawback is that the encoder-decoder framework utilizes a word-level strategy, which may lead to sub-optimal performance because the strategy and language functions are entangled together. 

Based on the above analysis, we elaborate three key issues in task-oriented dialog systems which will be discussed in detail shortly:

\begin{itemize}
	\item \textbf{Data Efficiency} Most neural approaches are data-hungry, requiring a large amount of data to fully train the model. However, in task-oriented dialog systems, the domain-specific data are often hard to collect and expensive to annotate. Therefore, the problem of low-resource learning is one of the major challenges.
	
	\item \textbf{Multi-turn Dynamics} Compared to open-domain dialog, the core feature of task-oriented dialog is its emphasis on goal-driven in multi-turn strategy.
	In each turn, the system action should be consistent with the dialog history and should guide the subsequent dialog to larger task reward. Nevertheless, the model-free RL methods which have shown superior performance on many tasks, can not be directly adopted to task-oriented dialog, due to the costly training environment and imperfect reward definition. Therefore, many solutions are proposed to tackle these problems in multi-turn interactive training for better policy learning, including model-based planning, reward estimation and end-to-end policy learning.
	
	\item \textbf{Ontology Integration} A task-oriented dialog system has to query the knowledge base (KB) to retrieve some entities for response generation. In pipeline methods, the KB query is mostly constructed according to DST results. Compared to pipeline models, the end-to-end approaches bypass modular models which requires fine-grained annotation and domain expertise. However, this simplification makes it hard to construct a query since there is no explicit state representation.
\end{itemize}

This paper is structured as follows: In Section 2, we introduce the recent advances of each component in pipeline methods and end-to-end approaches. In Section 3, we discuss recent work on task-oriented dialog evaluation, including automatic, simulated, and human evaluation methods. In Section 4, we survey some widely-used corpus for task-oriented dialog. In Section 5, we review the approaches proposed to address the above three challenges. Finally in Section 6, we conclude the paper and 
discuss future research trends.

\section{Modules and Approaches}

\begin{figure*}[!t]
	\centering
	\includegraphics[width=1.0\linewidth]{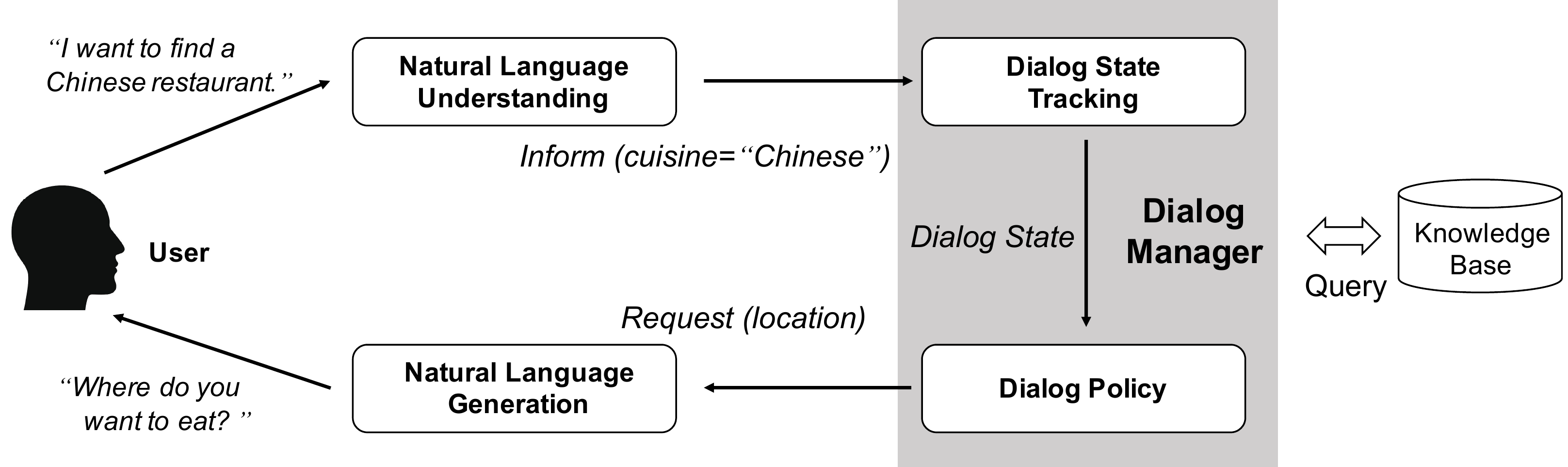}
	\caption{General framework of a pipeline task-oriented dialog system.}
	\label{fig:pipeline}
\end{figure*}


The architecture of task-oriented dialog systems can be roughly divided into two classes: pipeline and end-to-end approaches. In pipeline approaches, the model often consists of several components, including \textit{Natural Language Understanding} (NLU), \textit{Dialog State Tracking} (DST), \textit{Dialog Policy}, and \textit{Natural Language Generation} (NLG), which are combined in a pipeline manner as shown in Figure \ref{fig:pipeline}. The NLU, DST and NLG components are often trained individually before being aggregated together, while the dialog policy component is trained within the composed system. It is worth noting that although the NLU-DST-Policy-NLG framework is a typical configuration of the pipeline system, there are some other kinds of configurations. Recently, there are studies that merge some of the typical components, such as word-level DST and word-level policy, resulting in various pipeline configurations \cite{mrkvsic2017neural,wu2019transferable,zhao2019rethinking,chen2019semantically}.

In end-to-end approaches, dialog systems are trained in an end-to-end manner, without specifying each individual component. Commonly, the training process is formulated as generating a responding utterance given the dialog context and the backend knowledge base.

\subsection{Natural Language Understanding}

Given a user utterance, the natural language understanding (NLU) component maps the utterance to a structured semantic representation. A popular schema for semantic representation is the dialog act, which consists of intent and slot-values, as illustrated in Table \ref{tab:dialog_act}. The intent type is a high-level classification of an utterance, such as \textit{Query} and \textit{Inform}, which indicates the function of the utterance. Slot-value pairs are the task-specific semantic elements that are mentioned in the utterance. Note that both intent type and slot-value pairs are task-specific, which are related to the ontology and can be used to query knowledge base.

Based on the dialog act structure, the task of NLU can be further decomposed into two tasks: intent detection and slot-value extraction. The former is normally formulated as an intent classification task by taking the utterance as input, while the slot-value recognition task is often viewed as a sequence labeling problem:
\begin{ceqn}
\begin{align}
	p_{intent}(d|x_1, x_2, ..., x_n) \\
	p_{s-v}(y_1, y_2, ..., y_n|x_1, x_2, ..., x_n)
\end{align}
\end{ceqn}
where the $d$ indicates intent class and $y_1$ to $y_n$ are the labels of each token in the utterance $[x_1, x_2, ..., x_n]$ in which $x_i$ is a token and $n$ means the number of tokens.

Due to the strong ability of sequence modeling, RNN and its variants have been widely used in intent detection and slot-value extraction \cite{RNN4LU,LSTM4SLU,BLSTM4LU}.
These models used the hidden state of each token to predict the corresponding label $y_i$ and used the final hidden state to recognize the sentence intent $d$.
Other neural network structures such as recursive neural network \cite{recNN4LU} and CNN \cite{CNNtriCRF4LU} have also been explored. 
Conditional random field, which is frequently used by traditional sequence tagging models, was combined with RNN \cite{RNNCRF4LU} and CNN \cite{CNNtriCRF4LU} to improve the performance. 
Recently pre-training model BERT \cite{Devlin2019bert} has been another popular choice \cite{BERT4LU,Multi-lingual_BERT4LU}.

There are also some models strengthening the connection between intent classification and slot tagging. \cite{SlotGate} used an intent gate to direct the slot tagging process, while \cite{Attn4LU} applied attention mechanism to allow the interaction between word and sentence representations.




\begin{table*}[!t]
	\centering
	\caption{An example of dialog act for an utterance in the restaurant reservation domain.}
	\vspace{+0.3cm}
	\begin{tabular}{|c|c|c|c|c|c|}
		\hline
		{\bf Utterance} & \multicolumn{5}{c|}{How about a {\em British} restaurant in {\em north} part of town.}\\
		\hline
		{\bf Intent}&\multicolumn{5}{c|}{\em Query}\\
		\hline
		{\bf Slot Value}&\multicolumn{5}{c|}{Cuisine={\em British}, Location={\em North}}\\
		\hline
	\end{tabular}
	\label{tab:dialog_act}
\end{table*}

\subsection{Dialog State Tracking}

The dialog state tracker estimates the user's goal in each time step by taking the entire dialog context as input. The dialog state at time $t$ can be regarded as an abstracted representation of the previous turns until $t$.
Early works assumed some fixed sets of dialog state, and modeled the state transition during interaction as a Markov Decision Process (MDP). POMDP further assumes that the observation is partially observable, which makes it more robust in sophisticated situations \cite{young2013pomdp,young2006using,williams2005scaling,schatzmann2007agenda}.
Most recent works adopted belief state for dialog state representation, in which the state is composed of slot-value pairs that represent the user's goal. Therefore, this problem can be formulated as a multi-task classification task \cite{henderson2014word, mrkvsic2015multi,mrkvsic2017neural,lee2019sumbt}:
\begin{ceqn}
\begin{align}
	p_i(d_{i,t}|u_1, u_2, ..., u_t)
\end{align}
\end{ceqn}
where for each specific slot $i$, there is a tracker $p_i$. $u_t$ represents the utterance in turn $t$. The class of slot $i$ in the $t$-th turn is $d_{i,t}$.
However, this approach falls short when facing previously unseen values at run time.
Besides, there are also some works formulating the DST task as a reading comprehension task \cite{gao2019dialog,perez2020machine}.

In more recent methods, slots can be divided into two types: free-form and fixed vocabulary \cite{zhang2019find}. The former type does not assume a fixed vocabulary for the slot, which means the model cannot predict the values by classification. 
For free-form slot, one could generate the value directly \cite{wu2019transferable,ren2019scalable} or predict the span of the value in the utterance \cite{zhou2019multi,gao2019dialog}.
In generative methods, they often use a decoder to generate the value of a slot word by word from a large vocabulary. However, for rare words, this method can also fail since the vocabulary is limited. While for span-based methods, the model assumes that the value are shown in the context, and predicts the start and end position of that span.

\begin{figure*}[!t]
	\centering
	\includegraphics[width=0.7\linewidth]{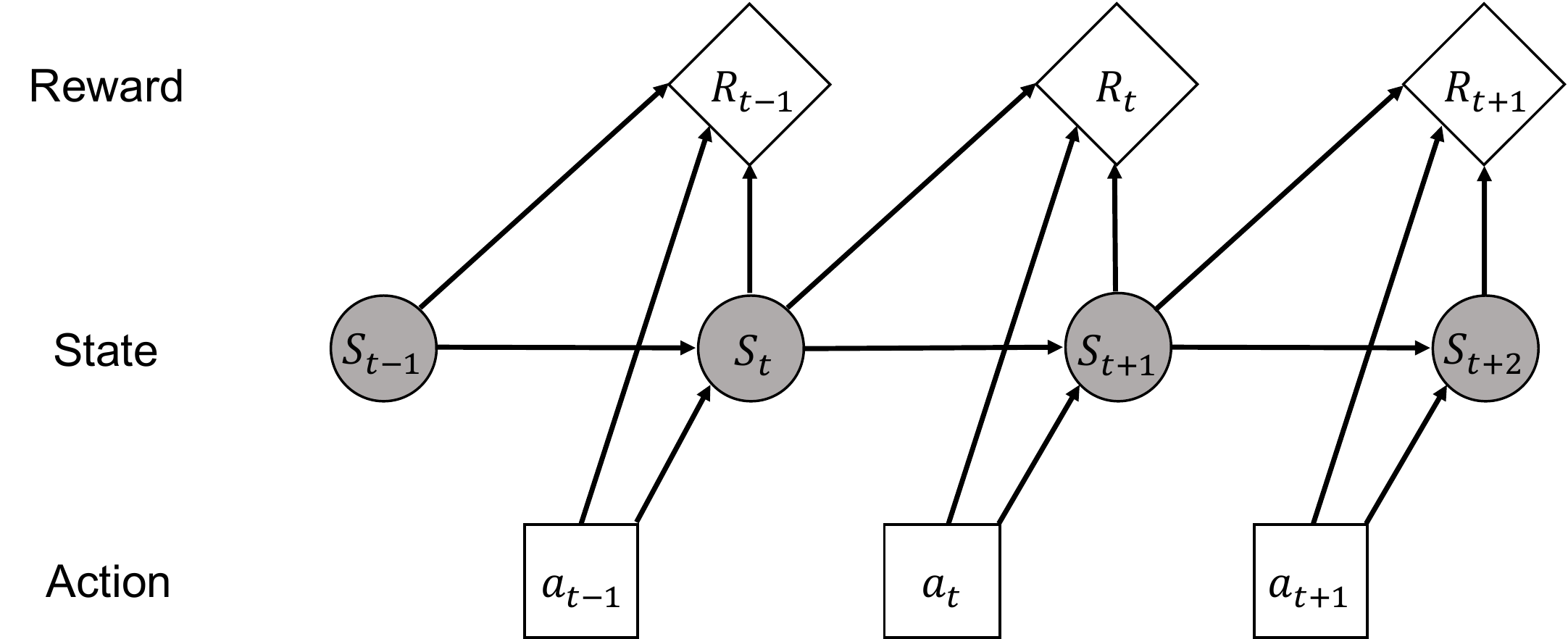}
	\caption{ Framework of Markov Decision Process \cite{poole2010artificial}. At time $t$, the system takes an action $a_t$, receiving a reward $R_t$ and transferring to a new state $S_{t+1}$.}
	\label{fig:mdp}
\end{figure*}

\subsection{Dialog Policy}

Conditioned on the dialog state, the dialog policy generates the next system action. Since the dialog acts in a session are generated sequentially, it is often formulated as a Markov Decision Process (MDP), which can be addressed by Reinforcement Learning (RL).
As illustrated in Figure \ref{fig:mdp}, at a specific time step $t$, the user takes an action $a_t$, receiving a reward $R_t$ and the state is updated to $S_t$.

A typical approach is to first train the dialog policy off-line through supervised learning or imitation learning based on a dialog corpus, and then fine-tune the model through RL with real users. Since real user dialogs are costly, user simulation techniques are introduced to provide affordable training dialogs.

Human conversation can be formulated as a Markov Decision Process (MDP): at each time step, the system transits from some state $s$ to a new state $s'$ by taking certain action $a$. Therefore, reinforcement learning is often applied to solve such an MDP problem in the dialog systems.

Model-free RL methods dominated the early studies of neural dialog policy by learning through interaction with real users, such as DQN and Policy Gradient methods \cite{mnih2015human, zhou2017end, lipton2018bbq}. For complex multi-domain dialogs, hierarchical RL models are introduced to first decide which is the domain of current turn and then select an action of that domain \cite{peng2017composite}.
Training a RL policy model requires a large amount of interactions .
One common solution is to use user simulators \cite{schatzmann2007agenda, li2016user, shi2019build}, which is another dialog system acting like a human user to provide training and evaluating environment. However, the user simulator is not able to fully mimic real human conversation behaviors, and its inductive bias may lead to sub-optimal models that perform poorly in real human conversation.
To alleviate these problems, model-based RL methods are proposed to model the environment, enabling planning for dialog policy learning \cite{peng2018deep,wu2019switch,su2018discriminative}.
In model-based RL approaches, the environment is modeled to simulate the dynamics of the conversation. Then in the RL training phase, the dialog policy is alternately trained through learning from real users and planning with the environment model. Some other works jointly train a system policy and a user policy simultaneously \cite{papangelis2019collaborative,takanobu2020multi}.

\subsection{Natural Language Generation}

Given the dialog act generated by the dialog policy, the natural language generation component maps the act to a natural language utterance, which is often modeled as a conditioned language generation task \cite{wen2015semantically}. The task takes dialog act as input and generates the natural language response.
To improve user experience, the generated utterance should (1) fully convey the semantics of a dialog act for task-completion, and (2) be natural, specific, and informative, analogous to human language.
Another problem is how to build a robust NLG with limited training data. Peng et al. \cite{peng2020few} proposed SC-GPT by first pre-training GPT with large-scale NLG corpus collected from existing publicly available dialog datasets, and then fine-tuning the model on target NLG tasks with few training instance.

\subsection{End-to-end Methods}
\begin{figure*}[!t]
	\centering
	\includegraphics[width=0.5\linewidth]{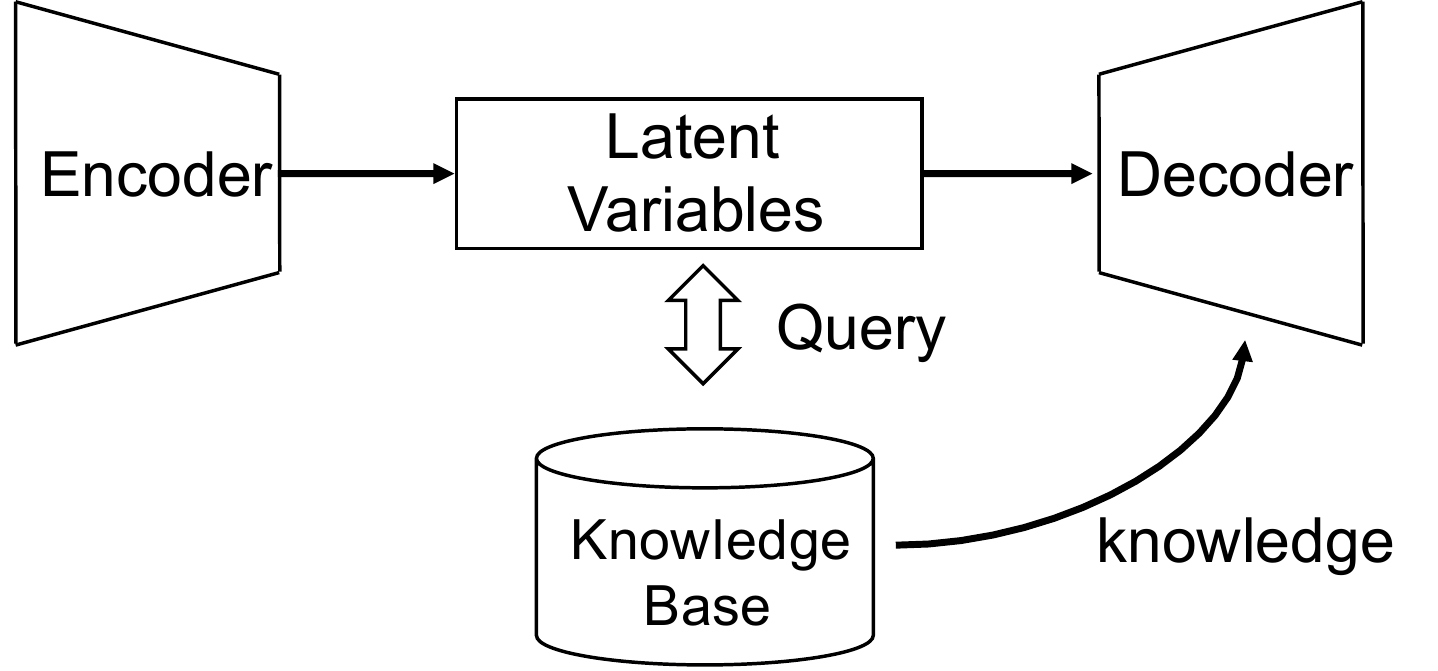}
	\caption{ Framework of end-to-end dialog systems. It first encodes natural language context to obtain some latent variables, which can be used for KB query. Then based on the latent variables and query results, the decoder generates a natural language response.}
	\label{fig:end2end}
\end{figure*}

Generally speaking, the components in a pipeline system are optimized separately. This modularized structure leads to complex model design, and the performance of each individual component does not necessarily translate to the advance of the whole system \cite{gao2018neural}.
The end-to-end approaches for task-oriented dialog systems are inspired by the researches on open-domain dialog systems, which use neural models to build the system in an end-to-end manner without modular design, as shown in Figure \ref{fig:end2end}. Most of these methods utilized sequence to sequence models as the infrastructural framework, which is end-to-end differentiable and can be optimized by gradient-based methods \cite{goodfellow2016deep}.

In most existing end-to-end approaches, the models are trained to maximize the prediction probability of response in the collected data.
Wen et al. \cite{wen2017network} proposed a modularized end-to-end model in which each component is modeled using neural networks, which makes the model end-to-end differentiable.
Bordes et al. \cite{bordes2017learning} formalized the task-oriented dialog as a reading comprehension task by regarding the dialog history as context, user utterance as the question, and system response as the answer. In this work, they utilized end-to-end memory networks for multi-turn inference. Madotto et al. \cite{madotto2018mem2seq} took a similar approach and further feed the knowledge base information into the memory networks. In \cite{eric2017key} a new memory network structure named key-value memory networks is introduced to extract relevant information from KB through key-value retrieval. Lei et al. \cite{lei2018sequicity} proposed a two-step seq2seq generation model which bypassed the structured dialog act representation, and only retain the dialog state representation. In their method, the model first encodes the dialog history and then generates a dialog state using LSTM and CopyNet. Given the state, the model then generates the final natural language response.

One major drawback of the above methods is that they often require large amounts of training data, which is expensive to obtain. Furthermore, they cannot fully explore the state-action space since the model only observes examples in the data. Therefore, reinforcement learning methods are introduced to mitigate these issues \cite{zhao2016towards, williams2017hybrid, dhingra2016towards, li2017end, liu2017iterative, lei2018sequicity}.
In \cite{zhao2016towards}, there is an end-to-end model that takes the natural language utterance as input and generates system dialog act as a response. In this method, there is no explicit state representation. Instead, they used LSTM to encode the dialog history into a state vector and then use DQN to select an action.
Williams et al. \cite{williams2017hybrid} proposed LSTM-based hybrid code networks (HCN), which supports self-defined software.

\section{Evaluation}

The evaluation of a dialog agent is crucial for the progress of task-oriented dialog systems. Most evaluation studies follow the PARADISE \cite{walker1997paradise} framework. It estimates the user satisfaction from two aspects. One is \textit{dialog cost} that measures the cost incurred in the dialog, such as the number of turns. The other one is \textit{task success} that evaluates whether the system successfully solves the user’s problem. The approaches to evaluate a task-oriented dialog system can be roughly grouped into the following three lines.

\subsection{Automatic Evaluation}

Automatic evaluation is widely advocated since it is quick, cheap, and objective. A bunch of well-defined automatic metrics have been designed for different components in the system. For language understanding, \textit{slot F1} and \textit{intent accuracy} are used. For dialog state tracking, the evaluation metrics include \textit{slot accuracy} and \textit{joint state accuracy} in general. For policy optimization, \textit{inform rate}, \textit{match rate} and \textit{task success rate} are used. For language generation, metrics such as \textit{BLEU} and \textit{perplexity} are applicable. Detailed definition of these metrics can be found in \cite{takanobu2020your}. All the models can be optimized against these metrics via supervised learning. However, each component is trained or evaluated separately in this way. Moreover, it assumes that the model would be fed with the ground truth from upstream modules or last dialog turn in the training process, but this assumption is invalid in real conversation.

\subsection{Simulated Evaluation}

In addition to training RL-based agents, a user simulator mimicking user behaviors in the task-oriented dialog also enables us to evaluate a trained dialog system. This is because, distinct from open-domain dialog systems, user goals in task-oriented dialog systems are somehow ``enumerable" so that it is feasible to exhaustively leverage domain expertise to build a user simulator, which can provide human-like conversational interaction for simulated evaluation. The metrics used in the simulated evaluation includes task success rate, dialog length, average rewards, etc. 

Simulated evaluation has been widely applied in the recently proposed dialog system platforms, such as PyDial \cite{ultes2017pydial} and ConvLab \cite{lee2019convlab,zhu2020convlab}.
The main advantage of simulated evaluation is that (1) the system can be evaluated in an end-to-end fashion; (2) multi-turn interaction is available during inference; (3) synthetic dialog data can be efficiently generated for evaluation at no cost. Similar to dialog policy optimization, the main challenge of employing simulated evaluation is to build a good user simulator that can mimic real user behaviors as much as possible. Meanwhile, how to evaluate the user simulator also remains an ongoing research direction \cite{pietquin2013survey}.

\subsection{Human Evaluation}

Simulated evaluation is efficient to evaluate the system performance with automatic, simulated interactions. Even though having a perfect user simulator, we still require human judgement for more complete evaluation on, e.g. covariate shift between the simulated environment and real conversation \cite{liu2018adversarial} and the quality of response generation \cite{wen2017network}, to assess real user satisfaction. Human evaluation metrics include task success rate, irrelevant turn rate, redundant turn rate, user satisfaction score, etc.

The researchers generally hire human users on the crowd-sourcing platform, and human evaluation can be conducted in the following two ways. One is \textit{indirect evaluation} that asking the annotators to read the simulated dialog between the dialog system and the user simulator, then rate the score \cite{shi2019build} or give their preference among different systems \cite{takanobu2019guided} according to each metric. The other one is \textit{direct evaluation} that the participants 
are asked to interact with the system to complete a certain task, give their ratings on the interaction experience. For example, \textit{language understanding} that evaluates whether the dialog agent understands user input, and \textit{response appropriateness} that evaluates whether the dialog response is appropriate during the conversation, are assessed in the DSTC8 competition \cite{li2020results}.


\section{Corpora}

A number of corpora with various domains and annotation granularity have been collected to facilitate the research on task-oriented dialog systems. Some datasets contain single-domain conversations \cite{henderson2014second,wen2017network,eric2017key,el2017frames}. With the increasing demands to handle various tasks in real-world applications, some large-scale multi-domain corpora \cite{budzianowski2018multiwoz,peskov2019multi,byrne2019taskmaster} have been collected recently. These datasets have higher language variation and task complexity. While most datasets are in English, 
Zhu et al. \cite{Zhu2020Crosswoz} propose the first large-scale Chinese task-oriented dataset with rich annotations to facilitate the research of
Chinese and cross-lingual dialog modeling. An incomplete survey on these dialog datasets is presented in Table \ref{tab:corpora}.

With respect to data annotation, the DSTC corpus \cite{williams2013dialog} provides the first common testbed and evaluation suite for dialog state tracking. DSTC2 \cite{henderson2014second} contains additional details on the ontology including a list of attributes termed \textit{informable slots} and \textit{requestable slots} for NLU tasks. User goals and a database of matching entities during the conversation are provided in some corpora \cite{wen2017network,budzianowski2018multiwoz,li2018microsoft} as well, which can be utilized for modeling multi-turn interactions. It is worth noting that, the schema of dialog state annotation is often different across these datasets. For example, \textit{search methods} representing user intents are included in DSTC2, and a schema listing the supported slots and intents along with their natural language descriptions is provided in SGD \cite{rastogi2020towards}.

There are mainly three modes in data collection. The first one is human-to-machine (H2M) where the data is collected via human users talking to a deployed machine-based system. The second mode is machine-to-machine (M2M) where two systems play the user and system roles respectively and interact to each other to generate conversations. Shah et al. \cite{shah2018bootstrapping} bootstrap the data collection process by first generating dialog templates at dialog act level using the M2M mode, and then converting these templates to natural language using crowd sourcing. The advantage of this method lies in that semantic annotations can be obtained automatically, which is thus cost-effective and error-resistant since translating templates to sentences is relatively simple for crowd-sourcing workers. However, the task complexity and language diversity is often restricted because the dialog simulation is performed using heuristic rules. 
The third mode is human-to-human (H2H), most following the Wizard-of-Oz (WoZ) paradigm \cite{kelley1984iterative} which collects real conversations between two crowd-sourcing workers who play a role of an agent (system) and a client (user) respectively. Each worker is given a task description about their goals and how they should act before the dialog is launched. While such a framework yields natural and diverse dialogs, it raises the difficulty of data annotation, especially when the annotation scheme is fine-grained. 

\begin{table*}[!htbp]
	\vspace*{0cm}
	\hspace*{-0.5cm}
	\centering
	\small
	\begin{tabular}{llllll}
		\toprule
		Name & Task & Method & Size & Statistics & Labels/Ontologies \\
		\midrule
		DSTC\cite{williams2013dialog} & Bus timetable & H2M & 15K & 14 turns/dialog & dialog states \\
		&&&&& user/system dialog acts \\
		DSTC2\cite{henderson2014second} & Restaurant booking & H2M & 3.2K & 14.49 turns/dialog & dialog states \\
		&&&& 8.54 tokens/turn & user/system dialog acts \\
		&&&&& database \\
		bAbI\cite{bordes2017learning} & Restaurant booking & M2M & 3K & 5 tasks & dialog level database \\
		CamRest\cite{wen2017network} & Restaurant booking & H2H & 676 & 7.4 turns/dialog & dialog states \\
		&&&&& user/system dialog acts \\
		&&&&& database \\
		WOZ\cite{mrkvsic2017neural} & Restaurant booking & H2H & 1.2K & 7.45 turns/dialog & dialog states \\
		&&&& 11.24 tokens/turn & user/system dialog acts \\
		&&&&& database \\
		KVReT\cite{eric2017key} & Car assistant & H2H & 3K & 5.25 turns/dialog & dialog states \\
		&&&& 8.02 tokens/turn & dialog level database \\
		Frames\cite{el2017frames} & Flight/Hotel booking & H2H & 1.4K & 14.6 turns/dialog & semantic frame \\
		&&&& 12.6 tokens/turn & user/system dialog acts \\
		&&&&& dialog level database \\
		SimD\cite{shah2018building} & Restaurant/Movie booking & M2M & 3K & 9.86 turns/dialog & dialog states \\
		&&&& 8.24 tokens/turn & user/system dialog acts \\
		AirD\cite{wei2018airdialogue} & Flight booking & M2M/H2H & 40K & 14.1 turns/dialog & dialog states\\
		&&&& 8.17 tokens/turn & database\\
		&&&&& context pairs \\
		MultiWOZ\cite{budzianowski2018multiwoz} & Multi-domain booking & H2H & 10K & 7 domains & dialog states \\
		&(Restaurant, Train, etc.)&&& 13.68 turns/dialog & system dialog acts \\
		&&&& 13.18 tokens/turn & database \\
		&&&&& user goals \\
		MDC\cite{li2018microsoft} & Movie/Restaurant/Taxi booking & H2H & 10K & 7.5 turns/dialog & user/system dialog acts \\
		&&&&& database \\
		&&&&& user goals \\
		CoSQL\cite{yu2019cosql} & Multi-domain booking & H2H & 3K & 138 domains & sql queries\\
		&(College, Music, etc.)&&& 10.36 turns/dialog & user dialog acts \\
		&&&& 11.34 tokens/turn & database \\
		&&&&& query goals \\
		Taskmaster\cite{byrne2019taskmaster} & Multi-domain booking & H2H/self & 13K & 6 domains & API calls and arguments \\
		& (Repair, Drinks, etc.) &&& 22.9 turns/dialog & \\
		&&&& 8.1 tokens/turn &  \\
		SGD\cite{rastogi2020towards} & Multi-domain booking & M2M & 23K & 17 domains & schema-guided dialog states\\
		&(Movie, Flight, etc.)&&& 20.44 turns/dialog & user/system dialog acts \\
		&&&& 9.75 tokens/turn & services \\
		CrossWOZ\cite{Zhu2020Crosswoz} & Multi-domain booking & H2H & 6K & 5 domains & user/system dialog states \\
		&(Attraction, Hotel, etc.)&&& 16.9 turns/dialog & user/system dialog acts \\
		&&&& 16.3 tokens/turn & database \\
		&&&&& user goals \\
		\bottomrule
	\end{tabular}
	\caption{Task-Oriented Dialog Corpora.}
	\label{tab:corpora}
\end{table*}

\section{Challenges}

\subsection{Data Efficiency}
Different from the research in open-domain dialog systems, data-driven approaches for task-oriented dialog systems often require fine-grained annotations to learn the dialog model in a specific domain, e.g., dialog act and state labels.
However, it is often difficult to obtain a large-scale annotated corpus in a specific domain since (1) collecting a domain-specific corpus is more difficult than in the open-domain setting due to its task-specific nature, and (2) annotating fine-grained labels requires a large amount of human resources which is very expensive and time-consuming. 
Therefore, we have to face the problem of improving the data efficiency of building task-oriented dialog systems, particularly in low-resource settings.

In this section, we review some recent approaches proposed to mitigate this issue. We first review transfer learning methods that acquire prior knowledge from large-scale data or adapt trained models from other tasks. Then, we introduce some unsupervised methods, which can directly learn in a low-resource setting with few annotations through heuristic rules. In addition, we also review recent efforts on building data-driven user simulators.

\subsubsection{Transfer Learning}
One major assumption of machine learning is that the training and test data  have the same distribution. However, in many real-world scenarios, this does not hold when we have only limited data in the target task but sufficient data in another task, with different data distributions.
Transfer learning is thus proposed to mitigate this problem by transferring knowledge from a source task to a target task.


The same issue often occurs in task-oriented dialog systems. For example, how can a dialog system for restaurant reservation be adapted to hotel booking when there are only limited data in the hotel domain? In such a situation, the two domains' ontologies are similar, sharing many dialog acts and slots. In this setting, transfer learning can considerably reduce the amount of target data required for this adaptation.
Besides domain-level transfer, knowledge can also be transferred in many other dimensions, including inter-person and cross-lingual transfer.
For domain transfer, Mrk{\v{s}}i{\'c} et al. \cite{mrkvsic2015multi} proposed to learn the dialog state tracking model through multi-task learning on multiple domain datasets to transfer knowledge across domains, which can improve the performance on all tasks. 
In \cite{ilievski2018goal}, Ilievski et al. proposed to directly transfer the parameters of shared slots from the source domain model to initialize the target model. Chen et al. \cite{chen2018policy} proposed to model dialog agent using several slot-dependent agents and a slot-independent agent to track the private and public slots across different domains. In \cite{rastogi2017scalable,ren2018towards}, the parameters of DST models are shared across domains and is independent of pre-defined value sets.
Therefore, the model is able to transfer to previously unseen domains.
Wu et al. \cite{wu2019transferable} further decoupled the domain and slot from the model parameters by taking domain and slot names as inputs to the DST model.

For transferring across disjoint tasks, Mo et al. \cite{mo2018cross} proposed to transfer the dialog policy model between domains by learning act and state transfer functions where there are no shared slots, which directly maps from the source feature space to the target space.

For personalized knowledge transfer, in \cite{mo2018personalizing}, a hybrid DQN policy is proposed to transfer knowledge across different customers, in which there is a general Q-function for all customers and a personalized one for each specific customer. When transferring to a new person, only a small amount of data is required to learn the personalized Q-function. Mo et al. \cite{mo2017fine} further transfers finer granularity phrase-level knowledge between different persons while keeping personal preferences of each user intact by designing a novel personal control gate within the RNN decoder framework.

The research on cross-lingual transfer is recently proposed. In \cite{schuster2018cross}, three cross-lingual methods are studied: (1) Translating the training data to the target language, (2) Pre-training cross-lingual embeddings and (3) Using a multilingual machine translation encoder to share knowledge for contextual word representations.

Model-agnostic methods are also proposed for transfer learning in dialog systems, which are mostly inspired by the Model-Agnostic Meta-Learning (MAML) framework \cite{finn2017model}. The MAML framework can learn a good initialized model by simulating the train-test procedure during learning. By applying such methods on NLG, the model can get better results in a low-resource setting and show better domain generalization \cite{mi2019meta, qian2019domain}. Madotto et al. \cite{madotto2019personalizing} further extended this method for personalized dialog systems by leveraging only a few dialogue samples collected from the target user without using the persona-specific descriptions.

Besides the above methods which transfer knowledge from a source model, there are also some works improving data efficiency by directly endowing the model or algorithms with prior knowledge to decrease data usage. For example, improved RL methods including ACER \cite{weisz2018sample} and BBQ-Networks \cite{lipton2018bbq} are proposed to enhance sample efficiency. In \cite{casanueva2018feudal}, the action selection process is decomposed into master action and primitive action selection, and the two actions are designed according to the domain ontology.

\subsubsection{Unsupervised Methods} 
A crucial issue in dialog policy learning is to estimate reward signal, which is hard to be obtained in real-world applications. Therefore, building a reward estimation model is necessary for dialog policy learning, particularly during RL training.
By regarding the dialog policy as a generator and the reward function as a discriminator, generative adversarial nets (GAN) can be employed to learn the reward function in an unsupervised manner.
Liu et al. \cite{liu2018adversarial} first used GAN to learn a binary reward function by discriminating simulated from real user dialogs. Xu et al. \cite{xu2019unsupervised} extended this idea for detecting dialog failure by using the predicted reward as an indicator of failure.
Su et al. \cite{su2016line} used another way for reward estimation using Gaussian Process. By modeling the uncertainty of predicted reward, the model can actively require human intervention on potential failure cases. In their experiment, the requirement for human intervention dramatically decreases with the reduction in the uncertainty of reward estimation, which remarkably relieve manual annotation.

In most studies, the ontology of a dialog system is built by human experts through elaborate domain engineering.
Another line of work is to assist the human experts in this process by learning the dialog structure from unlabeled corpus automatically. Shi et al. \cite{shi2019unsupervised} proposed to learn a finite state machine of the dialog procedure through a variational autoencoder (VAE) based approach.
They first pre-trained a VAE based dialog model using raw dialog data without intermediate annotations.
Then several dialog states can be discovered according to the latent variables. After that, a state transition diagram can be built by estimating the transition probabilities between states. There are also some works analyzing the structure of task-oriented to facilitate language understanding. Takanobu et al. \cite{takanobu2018weakly} proposed an RL method for topic segmentation and labeling in task-oriented dialog systems, which aims to detect topic boundaries among dialogue turns and assign topic labels to them.

Recently, pre-training methods show superior performance on many NLP tasks. In such approaches, extensive linguistic features can be transferred from large-scale unlabeled corpora using unsupervised pre-training tasks, such as mask language modeling (MLM) and next sentence prediction (NSP). Wolf et al. \cite{wolf2019transfertransfo} followed this way by first pre-training a transformer model on large-scale dialog data and then fine-tuning the model on a personalized dialog task with multi-task learning. Budzianowski et al. \cite{budzianowski2019hello} further explored this idea to task-oriented dialog without explicit standalone dialogue policy and generation modules. In this work, the belief state and database state are first converted to natural language text and then taken as input to the transformer decoder besides the context.

\subsubsection{User Simulation}
User simulation techniques alleviate the data-hungry issue of the RL-based dialog policy model by providing a theoretically infinite number of training interactions.
Early approaches focused on agenda-based user simulator (ABUS) \cite{schatzmann2007agenda}, which is commonly used in building task-oriented dialog systems. It maintains a stack-like structure representing the user's goal with some heuristics.
Building an agenda-based simulator requires the human expert to define the agenda and heuristics rules explicitly. However, for more complex tasks, it is not feasible to define an explicit agenda structure. Utterances from ABUS also lack linguistic variations of human dialogs, which may lead to suboptimal performance in real applications.

Recently, building user simulators in a data-driven fashion is proposed to alleviate the above issues.
Asri et al. \cite{el2016sequence} proposed a dialog act level seq2seq user simulation model that takes into account the dialog context.
Crook et al. \cite{crook2017sequence} presented another seq2seq model which takes as input natural language contexts and outputs natural language responses.
Kreyssig et al. \cite{kreyssig2018neural} introduced a neural user simulator (NUS), which mimics the user behavior of the corpus and generates word-level user responses.
Gur et al. \cite{gur2018user} proposed a hierarchical seq2seq user simulator (HUS) that first encodes the user goal and system turns, and then generates user dialog act. To generate more diverse user acts, they extended HUS to a variational version (VHUS) where the user turn is generated from an unobservable latent variable.

Another line of data-driven user simulators trains the simulator together with the target dialog system, which can be regarded as a multi-agent fashion. Liu et al. \cite{liu2017iterative} proposed to first train the dialog system and the simulator based on the dialog corpus through supervised learning, and then fine-tune both models by reinforcement learning. In this work, the system and the simulator are trained cooperatively, in which both agents share the same reward function.
The world model in the Deep Dyna-Q (DDQ) based dialog planning framework \cite{peng2018deep,su2018discriminative,wu2019switch}, which is updated during training, can also be regarded as a simulator. However, different from RL-based co-training, the world model in DDQ is updated through supervised learning using real experience.

The user simulators in the above methods are trained based on the human-agent dialog data. In addition to this, the human can also assist dialog policy learning by providing human demonstrations. Since the human guidance is expensive, Chang et al. \cite{chang2017affordable} compared various teaching schemes answering the question how and when to teach, to use the teaching budget more economically. Chen et al. \cite{chen2017agent} further proposed companion learning (CL) framework, which integrates rule-based policy and RL-based policy. Since the rule teacher is not as good as a human teacher, an uncertainty estimation is introduced to control the timing of consultation and learning.

\subsection{Multi-turn Dynamics}

Compared to open-domain dialog systems, one major feature of task-oriented dialog systems is the emphasis on multi-turn state-action dynamics, which is mainly related to dialog management (DST and Policy).
In open-domain dialog systems, the research focuses more on generating reasonable, consistent, and inter-personal responses to maximize user engagement \cite{huang2019challenges}. While for task-oriented dialog systems, although the above issues are still important, the completion of a specific task has been viewed as more critical.
Therefore, the research on dialog management, which is responsible for tracking the dialog state and flow of the conversation, acts as the pillar of a dialog system.

Human conversation can be broadly formulated as a Markov Decision Process (MDP): at each time step, the system transits from a certain state $s$ to a new state $s'$ by taking an action $a$. Therefore, reinforcement learning (RL) is often applied to solve such an MDP problem in the dialog systems.
Recent studies on the dialog management of task-oriented dialog systems are mainly focused on the following topics: 
(1) generative DST with value decoder for free-form slots, (2) dialog planning for better sample efficiency in policy learning, and (3) user goal estimation for predicting task success and user satisfaction.

\subsubsection{Generative DST}

Dialog state tracker plays a central role in task-oriented dialog systems by  tracking of a structured dialog state representation at each turn.
Most recent DST studies applied a word-level structure by taking natural language as input without NLU, which may avoid the errors propagated from the NLU component.
In early neural DST methods, \textit{belief state} is widely adopted for dialog state representation \cite{henderson2014word}, which maintains a distribution over all possible values for each slot. Therefore, early methods commonly formulated DST as a classification task \cite{henderson2013deep,mrkvsic2015multi,mrkvsic2017neural,zhang2019memory,zhang2019neural}.
Matthew et al. \cite{henderson2013deep} first proposed to use recurrent neural networks for word-level dialog state tracking by taking both natural language utterances and ASR scores as input features. Nikola et al. \cite{mrkvsic2017neural} proposed Neural Belief Tracker (NBT), a word-level dialog state tracker that directly reads from natural language utterances. NBT explicitly modeled the \textit{system request} and \textit{system confirm} operations through a gating mechanism.
However, these approaches can only deal with pre-defined slot values in the domain ontology vocabulary, which generally fall short in tracking unknown slot values during inference.

Zhong et al. \cite{zhong2018global} proposed to share parameters across slots and learn slot-specific features through a globally-locally self-attention mechanism, which can generalize to rare values with few training data. However, the rare values are still in-vocabulary words. Lei et al. \cite{lei2018sequicity} use a seq2seq model with two-stage CopyNet to generate belief spans and response at the same time, which obtain satisfactory results in OOV cases. 
In the first stage, a belief state CopyNet \cite{gu2016incorporating} takes the user utterance as input and generates a belief state span. Then in the second stage, based on the utterance and belief span, another CopyNet generates the response utterance.
Hu et al. \cite{xu2018end} proposed to use pointer network \cite{vinyals2015pointer} to extract unknown slot values, which showed superior performance over discriminative DST methods.
A more practical way is to use both extractive and discriminative methods to handle different type of slots \cite{zhang2019find}. For the free-form slots, such as \textit{hotel name} and \textit{departure date}, their value should be extracted from the utterance. While for those fixed-vocabulary slots like \textit{hotel star} and \textit{room category}, it is better to predict their value using a classifier.

Recently, some multi-domain datasets are proposed to promote the research in this direction \cite{budzianowski2018multiwoz,rastogi2020towards}. Compared to single-domain tasks, the DST in multi-domain scenario has to predict the domain of slot values. Wu et al. proposed TRADE  \cite{wu2019transferable}, a transferable multi-domain DST using seq2seq model with CopyNet \cite{gu2016incorporating} to predict values. The parameters are shared across domains, enabling zero-shot DST for unseen domains.
COMER \cite{ren2019scalable} further decreases the computation complexity of value decoding by first deciding the domain and slot, and then decoding the value.
In the decoding of the above methods, they first input the domain and slot names to the decoder, and then decode the value. If we take the domain and slot names as a form of ``question'', then the model can be regarded as a question answering model by taking the previous turns as context, domain-slot names as question and the value as answer. 
DSTQA \cite{zhou2019multi} added more elements into the ``question'' in addition to the names, such as the description text of domain and slots, values of fixed-vocabulary slots. They also encoded the intermediate dialog state graph using GNN to alleviate value decoding.
In \cite{chen2020schema}, Chen et al. proposed to use graph attention neural networks to model the relations across slots.

\subsubsection{Dialog Planning}

Model-free RL methods dominated the early studies of neural dialog policy by learning through interaction with real users \cite{mnih2015human,williams2017hybrid,cuayahuitl2017simpleds,dhingra2016towards}.
It is data-hungry, requiring a large amount of interactions to train a policy model effectively.
One common solution is to use user simulators \cite{schatzmann2007agenda,li2016user}. However, the user simulator is not able to fully mimic real human conversation behaviors, and its inductive bias may lead to sub-optimal models that perform poorly in real human conversation \cite{shi2019build}.

To alleviate these problems, model-based RL methods are proposed to model the environment, enabling planning for dialog policy learning.
In model-based RL approaches, the environment is modeled to simulate the dynamics of the conversation. Then in the RL training phase, the dialog policy is alternately trained through learning from real users and planning with the environment model \cite{Sutton1998}.
Peng et al. \cite{peng2018deep} proposed the Deep Dyna-Q (DDQ) framework, which first integrates model-based planning for task-oriented dialog systems.
In the DDQ framework, there is a \textit{world model}, which is trained on real user experience to capture the dynamics of the environment. The dialog policy is trained through both direct RL with real user and simulated RL with the world model. During training, the world model is also updated through supervised learning based on the increasing real experience.
The performance of the world model, which is crucial for policy learning, continues to improve during training. However, the ratio of real vs. simulated experience used for Q-learning is fixed in the original DDQ framework. Therefore, controlled planning \cite{su2018discriminative, wu2019switch} is proposed to mitigate this issue by dynamically adjusting the ratio of real to simulated experiences according to the performance of the world model.

The above methods for planning are referred to as \textit{background planning}, which improves the policy through training on simulated experience with the world model.
Another line of planning-based research is \textit{decision time planning}, which directly decides which action to take in a specific state $S_t$ based on some simulated experience. The simulated future steps can provide extra hints to facilitate decision making.
Planning used in this way can look much deeper than one-step ahead at decision time, which is common in human activities.
Taking the chess game for example, the players often conduct mental simulation by looking several steps ahead and then decide how to move the pieces.
Some works \cite{lewis2017deal,yarats2017hierarchical} introduced \textit{dialog rollout} planning into negotiation dialogs, in which the agent simulates complete dialogues in a specific state $S_t$ for several candidate responses to get their expected reward, and the response with the highest reward will be taken.
Instead of completing the dialogs and obtaining explicit rewards, Jiang et al. \cite{jiang2019towards} proposed to look only several limited steps ahead and use those steps as additional features for the policy model to alleviate decision making.


\subsubsection{User Goal Estimation}

In RL-based dialog models, the user's goal is crucial for policy learning.
Reward signal is an indirect reflect of the user's goal since it gives the user's satisfaction of a dialog.
One typical approach of reward function definition is to assign a large positive reward at the end of a successful session and a small negative penalty for each turn to encourage short conversations \cite{su2018reward}.
However, in real-world applications where the user goal is not available, this reward can not be estimated effectively.
Another problem is that the reward signals are not consistent when they are objectively calculated by predefined rules or subjectively judged by real users.
To alleviate the above issues, there are some studies that learn an independent reward function to provide a reliable supervision signal.

One method for reward estimation is off-line learning with annotated data \cite{yang2012predicting}. By taking the dialog utterances and intermediate annotations as input features, reward learning can be formulated as a supervised regression or classification task. The annotated reward can be obtained from either human annotation or user simulator. 
However, since the input feature space is complicated, a large amount of manual annotation is required, which is too costly.

To resolve the above problems, there is another line of work using on-line learning for reward estimation \cite{su2016line}. Reward estimation is often formulated as a Gaussian Process regression task, which can additionally provide an uncertainty measure of its estimation. In this setting, active learning is adopted to reduce the demand for estimating real reward signals in which the users are only asked to provide feedback when the uncertainty score exceeds a threshold. In other cases, when the estimation uncertainty is small, the estimated reward is utilized.

Instead of estimating the reward signals through annotated labels, Inverse RL (IRL) aims to recover the reward function by observing expert demonstrations. Adversarial learning is often adopted for dialog reward estimation through distinguishing simulated and real user dialogs \cite{liu2018adversarial,xu2019unsupervised,takanobu2019guided}.

\subsection{Ontology Integration}

One major issue in task-oriented dialog systems is to integrate the ontology of dialog into the dialog model, including domain schema and knowledge base.
In most previous methods, the domain schema is pre-defined and highly dependent on the corpus they use, e.g., the slots of restaurant domain contain address area, cuisine type, price range, etc.. As querying the database and retrieving the results are essential for a task-oriented dialog system to make decisions and produce appropriate responses, there are also many efforts to integrate external database or API calls recently.


However, ontology integration for task-oriented dialog models becomes more challenging, because of the large scale of task domains. Though the pre-defined ontology can be considered into model design, these approaches are coupled with domain schema and can not be easily transferred to a new task.
While increasing end-to-end models are proposed to alleviate the schema integration problem, it is not trivial to involve context information and knowledge base since, different from pipeline methods, there is no explicit dialog state representation to generate an explicit knowledge base query.

In this section, we introduce some recent advances on (1) dialog task schema integration and (2) knowledge base integration in task-oriented dialog models.


\subsubsection{Schema Integration}

Integrating the schema into a dialog model is critical for task-oriented dialog, since the value prediction of NLU and DST, and the action selection in Policy are highly dependent on the domain schema.
Early methods for NLU use classification for intent detection and sequence labeling for slot-value recognition. Therefore, the schema integration are mainly reflected in the model output layer design, e.g., one class for each intent.  Early DST methods utilized a similar way by giving a value probability distribution on the value vocabulary for each slot (also known as belief state). 
For NLG methods, the inputs are often structured dialog acts, and the encoder input structure is highly dependent on the representation structure.

The above schema integration methods basically couple the schema and model design together, yield poor scalability and domain generalization.
Recently, there are many methods trying to untie the domain scheme and model design. Convlab \cite{lee2019convlab} provides additional user dialog act annotation in the MultiWOZ \cite{budzianowski2018multiwoz} dataset to enable developers to apply NLU models in multi-domain, multi-intent settings. While most DST makes assumption that a slot in a belief state can only be mapped to a single value within a single turn, COMER \cite{ren2019scalable} extends the representation of dialog states with priority operator that considers the user's preference on slot values. Other works \cite{zhou2019multi,gulyaev2020goal} use question answering methods for DST by taking domain-slot descriptions as question. The values are regarded as answers, which are predicted by either extraction or generation based methods. In such methods, the model design is decoupled from the domain schema, and the schema of domain is represented by a natural language text, which makes it easy to transfer to new domain.
For NLG task, Peng et al. \cite{peng2020few} proposed SC-GPT, which treats the structured dialog act as a sequence of tokens, and feeds the sequence to the generation model.
By pre-training on large-scale da-response pairs, the model is able to capture the semantic structure of the sequence-based dialog act representation.
When extending to a new domain, only a small amount of training instances (~50) are required to achieve satisfactory performance. ZSDG \cite{zhao2018zero} learns a cross-domain embedding space that models the semantics of dialog responses so that it can instantly generalize to new situations with minimal data. Each service (domain) in SGD \cite{rastogi2020towards} provides a schema listing the supported slots and intents along with their natural language descriptions. These descriptions are used to obtain a semantic representation of these schema elements, making models applicable in a zero-shot setting.

\subsubsection{Knowledge Base Integration}
It is critical for a task-oriented dialog system to query the external knowledge base to get user's inquired information.
Early models or pipeline systems retrieved entries from the knowledge base by issuing a query based on the current dialog state during conversational interaction, which requires some manual effort.
Training an end-to-end dialog system without intermediate supervision will be more appealing due to the growing task complexities in task-oriented scenarios. However, different from pipeline approaches, there is no explicit structured dialog state representation in end-to-end methods. Therefore, the knowledge base interaction is conducted by using intermediate latent representation of the model and trained seamlessly through end-to-end training.

CopyNet and end-to-end memory networks are widely used for integrating knowledge into dialog systems through the attention mechanism.
The copy mechanism, however, can also be regarded as a memory network in which the encoder hidden states consist of the memory units. Eric et al. \cite{eric2017copy} presented a copy-based method depending on the latent neural embedding to attend to dialog history and copy relevant prior context for decoding. However, they can only generate entities that are mentioned in the context. More recent works use memory networks for prior dialog context and knowledge integration \cite{madotto2018mem2seq,wu2019global}. In such approaches, the dialog context and knowledge base are modeled into two memory nets. Then in the decoding phase, the decoder's hidden state is used to selectively query and copy information from those memory nets. A key problem in such a method is that dialog context and knowledge base are heterogeneous information from different sources. Lin et al. \cite{lin2019task} proposed to model heterogeneous information using historical information, which is stored in a context-aware memory, and the knowledge base tuples are stored in a context-free memory. In \cite{qin2019entity}, a two-step KB retrieval is proposed to improve the entities' consistency by first deciding the entity row and then selecting the most relevant KB column.

Besides fully end-to-end methods with few intermediate supervision, there are also some end-to-end models integrating domain prior knowledge into the model through dialog act and belief state annotations. Williams et al. \cite{williams2017hybrid} proposed hybrid code networks (HCNs), which combines an RNN with domain knowledge encoded as software and templates, which can considerably reduce the training data required. Wen et al. \cite{wen2017network} presented a modularized end-to-end task-oriented dialog model by combining several pre-trained components together, and then fine-tuning the model using RL in an end-to-end fashion. However, compared to seq2seq models, these methods are more like simplified versions of the pipeline model.

\section{Discussion and Future Trends}
In this paper, we review the recent advancements on task-oriented dialog systems and discuss three critical topics: data efficiency, multi-turn dynamics, and knowledge integration. 
In addition, we also review some recent progresses on task-oriented dialog evaluation and widely-used corpora.
Despite these topics, there are still some interesting and challenging problems. We conclude by discussing some future trends on task-oriented dialog systems:

\begin{itemize}
	\item \textbf{Pre-training Methods for Dialog Systems.} Data scarcity is a critical challenge for building task-oriented dialog systems. On the one hand, collecting sufficient data for a specific domain is time-consuming and expensive. On the other hand, the task-oriented dialog system is a composite NLP task, which is expected to learn syntax, reasoning, decision making, and language generation from not only off-line data but also on-line interaction with users, presenting more requests for fine-grained data annotation and model design.
	Recently, pre-trained models have shown superior performance on many NLP tasks \cite{peters2018deep,radford2018improving,Devlin2019bert}. In this vein, a base model is first pre-trained on large-scale corpora by some unsupervised pre-training tasks, such as masked language model and next sentence prediction. During the pre-training phase, the base model can capture implicit language knowledge, learning from the large-scale corpora. Using such implicit knowledge, the base model can fast adapt to a target task by simply fine-tuning on the data for the target task. This idea can also be applied to task-oriented dialog systems to transfer general natural language knowledge from large-scale corpora to a specific dialog task. Some early studies have shown the possibility of using pre-training models to model task-oriented dialogs \cite{mehri2019pretraining,mehri2019multi,wolf2019transfertransfo,budzianowski2019hello,peng2020few}.
	
	\item \textbf{Domain Adaptation.} Different from open-domain dialogs, the task-oriented conversations are based upon a well-defined domain ontology, which constrains the agent actions, slot values and knowledge base for a specific task. Therefore, to accomplish a task, the models of a dialog system are highly dependent on the domain ontology. However, in most existing studies, such ontology knowledge is hard-coded into the model. For example, the dialog act types, slot value vocabularies and even slot-based belief states are all embedded into the model. Such hard-coded ontology embedding raises two problems: (1) Human experts are required to analyze the task and integrate the domain ontology into the model design, which is a time-consuming process. (2) An existing model cannot be easily transferred to another task.
	Therefore, decoupling the domain ontology and the dialog model to obtain better adaptation performance is a critical issue. One ultimate goal is to achieve zero-shot domain adaptation, which can directly build a dialog system given an ontology without any training data, just like humans do. 
	
	\item \textbf{Robustness.} The robustness of deep neural models has been a challenging problem since existing neural models are vulnerable to simple input perturbation. As for task-oriented dialog systems, robustness is also a critical issue, which mainly comes from two aspects: (1) On the one hand, the task-oriented dialogs are highly dependent on the domain ontology. Therefore, in many studies, the training data are constrained to only reasonable instances with few noises. However, models trained in such an ad hoc way often fall short in real applications where there are many out-of-domain or out-of-distribution inputs \cite{zheng2019out}, such as previously unseen slot values. A robust dialog system should be able to handle noises and previously unseen inputs after deployment. (2) On the other hand, the decision making of a neural dialog policy model is not controllable, which is trained through off-line imitation learning and on-line RL. The robustness of decision making is rather important for its performance, especially for some special applications which have a low tolerance for mistakes, such as in medical and military areas. 
	Therefore, improving the robustness of neural dialog models is an important issue. One possible approach is to combine robust rule-based methods with neural models, such as Neural Symbolic Machine \cite{liang2016neural,segler2017neural}, which may make the models not only more robust but also more explainable.
	
	\item \textbf{End-to-end Modeling.} Compared to pipeline approaches, end-to-end dialog system modeling is gaining more and more attention in recent years. The end-to-end model can be trained more easily without explicit modeling of dialog state and policy. However, existing end-to-end methods still require some intermediate supervision to boost the model performance. For example, in \cite{wen2017network}, a modular-based end-to-end framework is proposed by combining pre-trained components together and then fine-tuning all the components using RL in an end-to-end fashion, which still requires intermediate supervision such as dialog act and belief state at the pre-training phase. In \cite{lei2018sequicity}, although a seq-to-seq framework is proposed to avoid component barriers, intermediate output named \textit{belief spans} is still retained for explicit belief state modeling. Therefore, the problem of modeling task-oriented dialog in a fully end-to-end fashion, without intermediate supervision and can seamlessly interact with the knowledge base, is still an open problem.
\end{itemize}



\Acknowledgements{This work was supported by the National Science Foundation of China (Grant No.61936010/61876096) and the National Key R\&D Program of China (Grant No. 2018YFC0830200).}



\bibliographystyle{unsrt}
\bibliography{main}


%
%
%
%
%
%
%
%

\end{multicols}

\end{document}